\documentclass[12pt]{elsarticle}

\usepackage{url}
\usepackage{xcolor}
\usepackage[dvips]{epsfig}
\usepackage{lettrine}
\usepackage{float}
\usepackage{amsmath, amssymb}
\usepackage[tight,footnotesize]{subfigure}
\usepackage{multirow}
\usepackage{dsfont}
\usepackage{float}
\usepackage{algorithm}
\usepackage{algorithmic}
\usepackage{booktabs}
\usepackage{amsthm}
\usepackage{amssymb}
\usepackage[colorlinks, linkcolor=black, urlcolor=blue,anchorcolor=black, citecolor=blue]{hyperref}

\newtheorem{lemma}{Lemma}

\newproof{pf}{Proof}
\journal{Knowledge based Systems}

\begin{document}

\begin{frontmatter}



\title{Fast Low Rank Representation based Spatial Pyramid Matching for Image Classification}


\author [i2r]{Xi Peng}\ead{pangsaai@gmail.com}
\author [i2r]{Rui Yan}\ead{ryan@i2r.a-star.edu.sg}
\author [i2r]{Bo Zhao}\ead{zhaob@i2r.a-star.edu.sg}
\author [i2r]{Huajin Tang\corref{cor1}}\ead{htang@i2r.a-star.edu.sg}
\author [scu]{Zhang Yi}\ead{zhangyi@scu.edu.cn}
\address [i2r]{Institute for Infocomm Research,
Agency for Science, Technology and Research (A*STAR) Singapore 138632}
\address [scu]{Machine Intelligence Laboratory, College of Computer Science, Sichuan University, Chengdu, 610065, China.}
\cortext[cor1]{Corresponding author}
%\cortext[cor2]{Principal Corresponding author}

\begin{abstract}
Spatial Pyramid Matching (SPM) and its variants have achieved a lot of success in image classification. The main difference among them is their encoding schemes. For example, ScSPM incorporates Sparse Code (SC) instead of Vector Quantization (VQ) into the framework of SPM. Although the methods achieve a higher recognition rate than the traditional SPM, they consume more time to encode the local descriptors extracted from the image. In this paper, we propose using Low Rank Representation (LRR) to encode the descriptors under the framework of SPM. Different from SC, LRR considers the group effect among data points instead of sparsity. Benefiting from this property, the proposed method (i.e., LrrSPM) can offer a better performance. To further improve the generalizability and robustness, we reformulate the rank-minimization problem as a truncated projection problem. Extensive experimental studies show that LrrSPM is more efficient than its counterparts (e.g., ScSPM) while achieving competitive recognition rates on nine image data sets.
\end{abstract}


\end{frontmatter}

\section{Introduction}
\label{sec1}

Image classification system automatically assigns an unknown image to a category according to its visual content, which has been a major research direction in computer vision and pattern recognition. Image classification has two major challenges. First, each image may contain multiple objects with similar low level features, it is thus hard to accurately categorize the image using the global statistical information such as color or texture histograms. Second, a medium-sized grayscale image (e.g., $1024\times 800$) corresponds to a vector with dimensionality of $819,200$, this brings up the scalability issue with image classification techniques. 

To address these problems, numerous impressive approaches~\cite{Duchenne2011,Du201551,Mendoza2012,Lu2013,Yang2015Coupled} have been proposed in the past decade, among which one of the most popular methods is Bag-of-Features (BOF) or called Bag-of-Words (BOW). BOW originates from document analysis~\cite{Joachims1996,Blei2003}. It models each document as the joint probability distribution of a collection of words. \cite{Sivic2003,Csurka2004,Fei2005} incorporated the insights of BOW into image analysis by treating each image as a collection of unordered appearance descriptors extracted from local patches. Each descriptor is quantized into a discrete ``visual words'' corresponding to a given codebook (i.e., dictionary), and then the compact histogram representation is calculated for semantic image classification. 

The huge success of BOF has inspired a lot of works~\cite{Grauman2005,Bolovinou2013}. In particular, Lazebnik et\ al.~\cite{Lazebnik2006} proposed Spatial Pyramid Matching (SPM) which divides each image into $2^{l}\times 2^{l}$ blocks in different scales $l=0,1,2$, then computes the histograms of local features inside each block, and finally concatenates all	 histograms to represent the image. Most state-of-the-art systems such as~\cite{Yu2009,Wang2010,Yu2011,Gao2013TIP, Zhou2013} are implemented under the framework of SPM and have achieved impressive performance on a range of image classification benchmarks like Columbia University Image Library-100 (COIL100)~\cite{COIL100} and Caltech101~\cite{Fei2006}. Moreover, SPM has been extensively studied for solving other image processing problems, e.g., image matching~\cite{Hur2015}, fine-grained image categorization~\cite{Zhang2015fine}. It has also been incorporated into deep learning to make deep convolutional neural networks (CNN)~\cite{He2014spatial} handling arbitrary sized images possible. To obtain a good performance, SPM and its extensions have to pass the obtained representation to a Support Vector Machine classifier (SVM) with nonlinear Mercer kernels. This brings up the scalability issue with SPMs in practice. 

Although SPM has achieved state-of-the-art recognition rates on a range of databases, its computational complexity is very high. To speed up SPM, Yang et\ al.~\cite{Yang2009} proposed using Sparse Code (SC) instead of Vector Quantization (VQ) to encode each Scale-Invariant Feature Transform (SIFT) descriptor~\cite{Lowe2004} over a codebook. Benefiting from the good performance of sparse code, Yang's method (namely ScSPM) with linear SVM obtains a higher classification accuracy, while using  less time for training and testing. 

The success of ScSPM could be attributed to that SC can capture the manifold structure of data sets. However, SC encodes each data point independently without considering the grouping effect among data points. Moreover, the computational complexity of SC is proportional to the cube of the size of codebook (denoted by $n$). Therefore, it is a daunting task to perform ScSPM when $n$ is larger than $10,000$. To solve these two problems, this paper proposes using Low Rank Representation (LRR) rather than SC to hierarchically encode each SIFT descriptor.

\begin{figure*}[!t]
\centering{
\subfigure[]{\label{fig1.a}\includegraphics[width=0.48\textwidth]{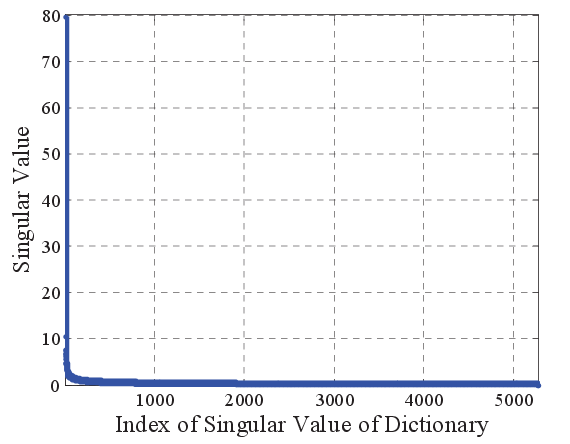}}\hspace{1mm}
\subfigure[]{\label{fig1.b}\includegraphics[width=0.48\textwidth]{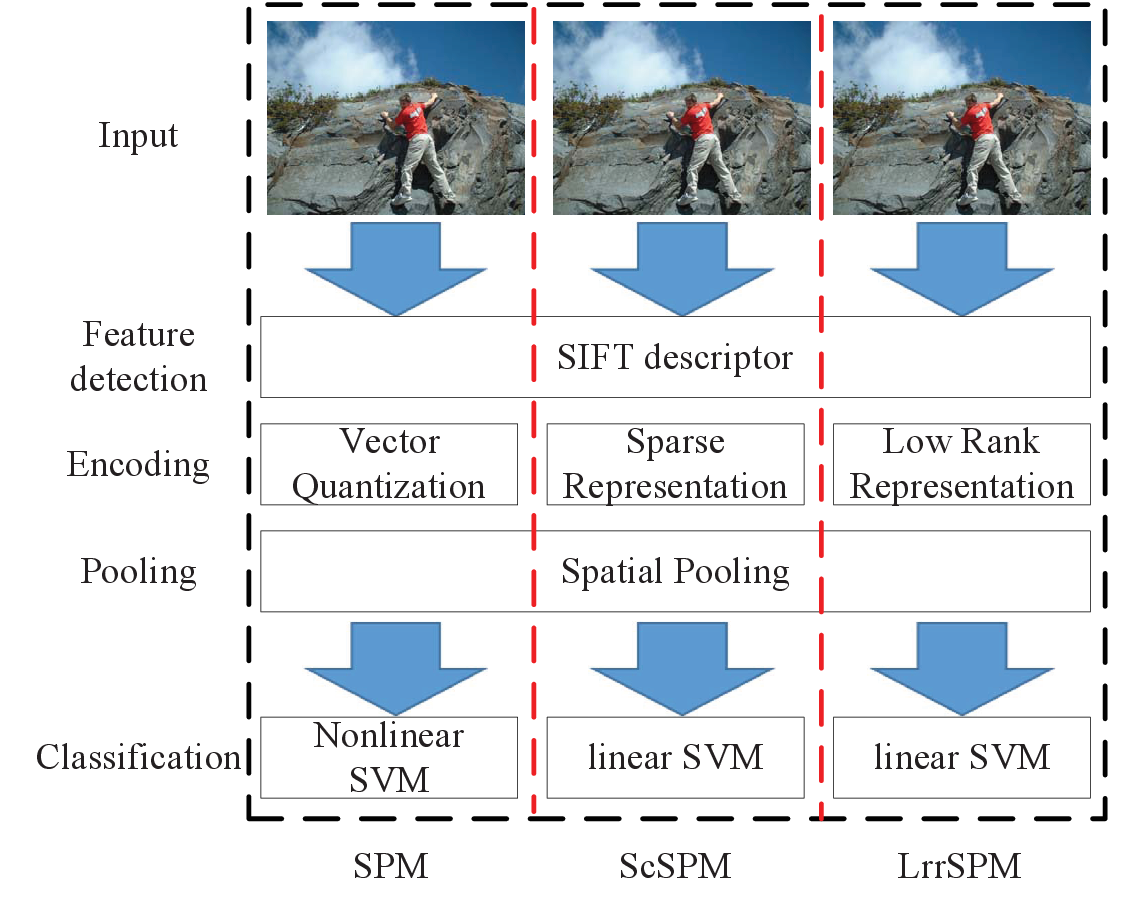}}
}
\caption{ (a) Singular values of a given codebook. The codebook consists of 5,120 basis. It shows that most energy concentrates on the top singular values. (b) Schematic comparison of the original SPM, ScSPM and the proposed LrrSPM. }
\label{fig1}
\end{figure*}

To show our motivation, i.e., the collections of descriptor and representation are low rank, we carry out k-means clustering algorithm over the SIFT descriptors of the Caltech101 database~\cite{Fei2006} and obtain a codebook $\mathbf{D}$ consisting of 5,120 cluster centers. By performing Singular Value Decomposition (SVD) over the codebook shown in~\figurename~\ref{fig1.a}, one can see that most energy (over 98\%) concentrates on the top 2\% singular values. In other words, the data space spanned by the codebook is low rank. For a testing data set $\mathbf{X}\in span(\mathbf{D})$, its representation can be calculated by $\mathbf{X}=\mathbf{D}\mathbf{C}$. Since $\mathbf{X}$ and $\mathbf{D}$ are low rank, then $\mathbf{C}$ must be low rank. This observation motivates us to develop an novel SPM method, namely Low Rank Representation based Spatial Pyramid Matching (LrrSPM).~\figurename~\ref{fig1} illustrates a schematic comparison of the original SPM, ScSPM, and LrrSPM. It should be pointed out that, SPM, ScSPM, and LrrSPM are three basic models which do not incorporate the label information, kernel function learning, and multiple descriptors learning into their encoding schemes. The major difference among them is that both SPM and ScSPM perform encoding in the vector space, whereas LrrSPM calculates the representation in the matrix space.

The contributions of the paper are summarized as follows: 1) Different from the existing LRR methods~\cite{Liu2013,Favaro2011,Xiao2014}, the proposed LrrSPM is a multiple-scale model which integrates more discriminative information compared to the traditional LRR. 2) Most existing LRR methods are proposed for clustering, which cannot be used for classification directly. In this paper, we fill this gap based on our new mathematical formulation. 3) Our LrrSPM has a closed form solution and can be calculated very fast. After the dictionary is learnt from the training data, LrrSPM computes the representation of testing data by simply projecting each testing datum into another space. Extensive experimental results show that LrrSPM achieves competitive results on nine image databases and is $25-50$ times faster than ScSPM. 

The rest of the paper is organized as follows:
Section~\ref{sec2} provides a brief review on two classic image classification methods, i.e., SPM~\cite{Lazebnik2006} and ScSPM~\cite{Yang2009}.
Section~\ref{sec3} presents our method (i.e., LrrSPM) which uses multiple-scale low rank representation to represent each image. 
Section~\ref{sec4} carries out some experiments using nine image data sets and several popular approaches.  
Finally, Section~\ref{sec5} concludes this work.

\begin{table}
\begin{center}
\begin{scriptsize}
   \caption{Some used mathematic notations.}
    \label{tab1}
    \begin{tabular}{ll}
    \toprule
    Notation & Definition\\
    \midrule
    $n$ & the number of descriptors (features) \\
    $l$  & the scale or resolution of a given image\\
%    $N=\sum(2^{l}\times 2^{l})$ & the number of blocks or subregions\\
    $m$ & the dimensionality of the descriptors\\
   $s$ & the number of subjects\\
   $k$ & the size of codebook\\
   $r$ & the rank of a given matrix\\
   $\mathbf{y}$ & an image\\
   $\mathbf{X}=[\mathbf{x}_1, \mathbf{x}_2, \ldots, \mathbf{x}_{n}]$ & a set of features \\
  $\mathbf{D}=[\mathbf{d}_1, \mathbf{d}_2, \ldots, \mathbf{d}_{k}]$ & codebook\\
$\mathbf{C}=[\mathbf{c}_1, \mathbf{c}_2, \ldots, \mathbf{c}_{n}]$ & the representation of $\mathbf{X}$ over $\mathbf{D}$\\
    \bottomrule
    \end{tabular}
\end{scriptsize}
\end{center}
\end{table}

\textbf{Notations:}  Lower-case bold letters represent column vectors and upper-case bold ones denote matrices. $\mathbf{A}^T$ and $\mathbf{A}^{-1}$ denote the transpose and pseudo-inverse of the matrix $\mathbf{A}$, respectively. $\mathbf{I}$ denotes the identity matrix. \tablename~\ref{tab1} summarizes some notations used throughout the paper.

\section{Related works}
\label{sec2}

In this section, we mainly introduce SPM and ScSPM which employ two basic encoding schemes, i.e., vector quantization and sparse code. To the best of our knowledge, most of other SPM based methods can be regarded as the extensions of them, e.g., the method proposed in~\cite{Gao2013TIP} is a kernel version of ScSPM. 

Let $\mathbf{X}\in\mathds{R}^{m\times n}$ be a collection of the descriptors and each column vector of $\mathbf{X}$ represents a feature vector $\mathbf{x}_{i}\in \mathds{R}^{m}$, SPM)~\cite{Lazebnik2006} applies VQ to encode $\mathbf{x}_{i}$ via 

\begin{equation}
\label{eq2.1}
\mathop{\min}_{\mathbf{C}, \mathbf{D}}{\sum_{i=1}^{n}\|\mathbf{x}_{i}-\mathbf{D}\mathbf{c}_{i}\|_{2}^{2}} \hspace{3mm} \mathrm{s.t.}\hspace{1mm}Card(\mathbf{c}_{i})=1,
\end{equation}
where $\|\cdot\|_2$ denotes $\ell_2$-norm, $\mathbf{c}_{i}\in \mathds{R}^{k}$ is the representation or called the cluster assignment of $\mathbf{x}_{i}$, the constraint $Card(\mathbf{c}_{i})=1$ guarantees that only one entry of $\mathbf{c}_{i}$ is with value of one and the rest are zeroes, and $\mathbf{D}\in \mathds{R}^{m\times k}$  denotes the codebook. 

In the training phase, $\mathbf{D}$ and $\mathbf{C}$ are iteratively solved, and VQ is equivalent to the classic k-means clustering algorithm which aims to

\begin{equation}
\label{eq2.2}
\mathop{\min}_{\mathbf{D}}\sum_{i=1}^{n}\mathop{\sum}_{j=1}^{k}\min\|\mathbf{x}_{i}-\mathbf{d}_{j}\|_{2}^{2},
\end{equation}
where $\mathbf{D}$ consists of $k$ cluster centers identified from $\mathbf{X}$.

In the testing phase, each $\mathbf{x}_{i}\in \mathbf{X}$ is actually assigned to the nearest $\mathbf{d}_{j}\in \mathbf{D}$. Since each $\mathbf{c}_{i}$ has only one nonzero element, it discards a lot of information for $\mathbf{x}_{i}$ (so-called hard coding problem). To solve this problem, Yang et\ al.~\cite{Yang2009} proposed ScSPM which uses sparse representation to represent  $\mathbf{x}_{i}$ via

\begin{equation}
\label{eq2.3}
\mathop{\min}_{\mathbf{c}_{i}} \|\mathbf{x}_{i}-\mathbf{D}\mathbf{c}_{i}\|_2^{2}+\lambda\|\mathbf{c}_{i}\|_{1},
\end{equation}
where $\|\cdot\|_{1}$ denotes $\ell_1$-norm which sums the absolute values of a vector, and $\lambda>0$ is the sparsity parameter. 

The advantage of ScSPM is that the sparse representation $\mathbf{c}_{i}$ has a small number of nonzero entries and it can represent $\mathbf{x}_{i}$ better with less reconstruction errors. Extensive studies~\cite{Wang2010,Yang2009} have shown that ScSPM with linear SVM is superior to the original SPM with nonlinear SVM. The disadvantage of ScSPM is that each data point $\mathbf{x}_{i}$ is encoded independently, and thus the sparse representation $\mathbf{c}_{i}$ cannot reflect the class structure. Moreover, the computational complexity of sparse coding is very high so that  any medium-sized data set will bring up scalability issue with ScSPM. Motivated by SPM and ScSPM, a lot of works have been proposed, e.g., nonlinear extensions~\cite{Gao2013TIP}, supervised extensions~\cite{Zhou2013}, and multiple descriptors fusing~\cite{Yang2009Group}.

\section{Fast Low Rank Representation Learning for Spatial Pyramid Matching}
\label{sec3}

In this section, we introduce LrrSPM in three steps. First, we give the basic formulation of LrrSPM which requires solving a rank-minimization problem. Moreover, we theoretically show that the representation is also low rank if the data space spanned by the codebook is low rank. This provides a  theoretical foundation for our method. Second, we further improve the generalization ability and robustness of the basic model by adopting the regularization technique and recent theoretical development in robustness learning. Finally, a real-world example is given to show the effectiveness of the obtained representation.

LRR  can capture the relations among different subjects, which has been widely studied in image clustering~\cite{Xiao2014}, semi-supervised learning~\cite{Yang2014}, and dimension reduction~\cite{Liu2011}. In this paper, we introduce LRR into SPM to hierarchically encode each local descriptor. Note that, it is nontrivial to incorporate LRR into the framework of SPM due to the following reasons. 1) the traditional LRR are generally used for clustering, which cannot be directly used for classification. In the context of classification, we need to reformulate the objective function that must lead to a different optimization problem. 2) To improve the robustness, the traditional LRR enforces $\ell_1$-norm over the possible errors, which results a very high computational complexity. In this paper, we do not take this error-removal strategy but perform truncated operator into the projection space to eliminate the effect of errors.

\begin{figure}[t]
\includegraphics[width=0.96\textwidth]{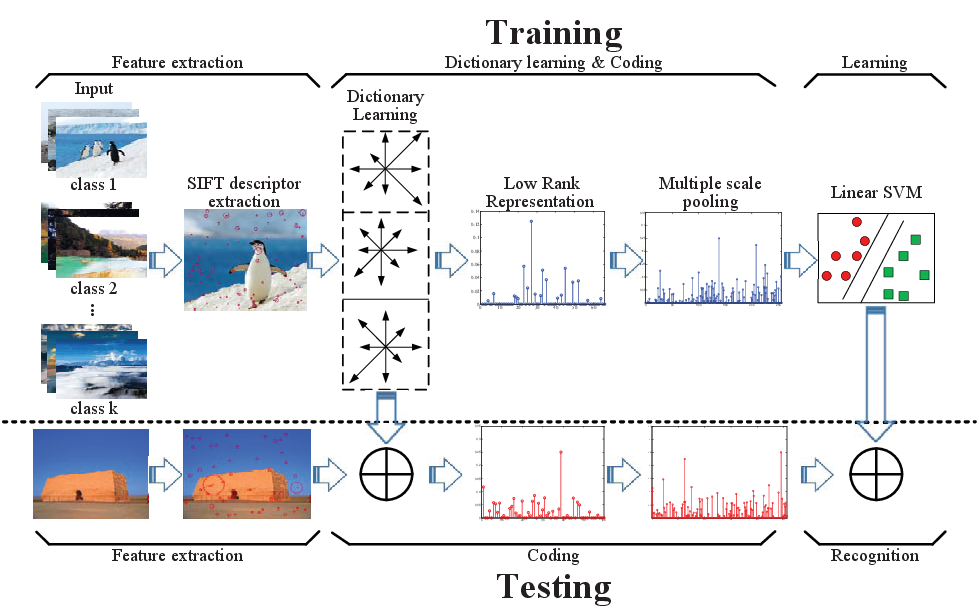}
\caption{Flow chart of the proposed algorithm.}
\label{fig2}
\end{figure}

\figurename~\ref{fig2} provides a flow chart of the proposed algorithm. Different from the existing LRR methods, we use the training data rather than all samples as codebook. We aim at solving

\begin{equation}
\label{eq3.1}
\mathop{\min}_{\mathbf{C}} \hspace{1mm} rank(\mathbf{C}) \hspace{3mm} \mathrm{s.t.}\hspace{1mm} \mathbf{X}=\mathbf{DC},
\end{equation}
where $\mathbf{X}\in \mathds{R}^{m\times n}$ denotes the collection of SIFT descriptors, $\mathbf{C}^{n\times k}$ denotes the representation of $\mathbf{X}$ over the codebook $\mathbf{D}^{m\times k}$, and $\mathbf{D}^{m\times k}$ generally consists of $k$ cluster centers. 

Note that, LrrSPM (i.e., eq.\ref{eq3.1}) enforces rank minimization operator over the representation matrix, which is significantly different from the standard SPM (i.e., eq.\ref{eq2.1}) and ScSPM (i.e., eq.\ref{eq2.3}). LrrSPM exploits the grouping effect of  data points in 2-dimensional space, whereas SPM and ScSPM obtain the representation in 1-dimensional space (i.e., vector).

The optimal solution to eq.\ref{eq3.1} is given by 

\begin{equation}
	\label{eq3.2}
	\mathbf{C}^{\ast}=\mathbf{D}^{-1}\mathbf{X},
\end{equation}
where $\mathbf{D}^{-1}$ denotes the inverse of $\mathbf{D}$. Note that, one always calculates the pseudo-inverse of $\mathbf{D}$ in practice, denoted by $\mathbf{C}^{\ast}=\mathbf{D}^{\dag}\mathbf{X}$. 

Based on the above results, we have
\begin{align}
	\label{eq3.3}
	\mathrm{rank}(\mathbf{C}^{\ast})&\le \min\{\mathrm{rank}(\mathbf{D}^{\dag}), \mathrm{rank}(\mathbf{X})\} \notag\\
	&= \min\{\mathrm{rank}(\mathbf{D}), \mathrm{rank}(\mathbf{X})\}.
\end{align}
This shows how the rank of $\mathbf{D}$ affects that of $\mathbf{C}$. Moreover, it also verifies our motivation once again, i.e., $\mathbf{C}$ must be low rank when the dictionary $\mathbf{D}$ is low rank.

To avoid overfitting, we further incorporate the regularization technique and obtain the following solution:

\begin{equation}
	\label{eq3.3}
	\mathbf{C}^{\ast}=(\mathbf{D}^{T}\mathbf{D}+\lambda\mathbf{I})^{-1}\mathbf{D}^{T}\mathbf{X},
\end{equation}
where $\lambda\ge 0$ is the regularization parameter and $\mathbf{I}$ denotes the identity matrix. Note that, nuclear-norm based representation is actually equivalent to the frobenius-norm based representation under some conditions, please refer to~\cite{Peng2015Connections} for more theoretical details.
 
In practice, $\mathbf{D}$ probably contains the errors such as noise, and thus the obtained representation may be sensitive to various corruptions. To achieve robust results, we recently proved that \textit{the trivial coefficients (i.e., small coefficients) always correspond to the representation over errors} in $\ell_2$-norm based projection space, i.e.,  

\begin{lemma}[\cite{peng2015robust}]
\label{lem1}
For any nonzero data point $\mathbf{x}$ in the
subspace $\mathcal{S}_{\mathbf{D}_{x}}$ except the intersection
between $\mathcal{S}_{\mathbf{D}_{x}}$ and
$\mathcal{S}_{\mathbf{D}_{-x}}$, i.e., $\mathbf{x}\in
\{\mathcal{S}|\mathcal{S}=\mathcal{S}_{\mathbf{D}_{x}}\backslash
\mathcal{S}_{\mathbf{D}_{-x}}\}$, the optimal solution of

\begin{equation}
\label{lem1eq1}
\min\hspace{1mm}\|\mathbf{c}\|_{2} \hspace{3mm}
\mathrm{s.t.}\hspace{1mm} \mathbf{x}=\mathbf{D}\mathbf{c},
\end{equation}
 over $\mathbf{D}$ is given by
$\mathbf{c}^{\ast}$ which is partitioned according to the sets
$\mathbf{D}_{x}$ and $\mathbf{D}_{-x}$, i.e.,
$\mathbf{c}^{\ast}=\begin{bmatrix}\mathbf{c}_{x}^{\ast}\\\mathbf{c}_{-x}^{\ast}
\end{bmatrix}$. Thus, we must have
$[\mathbf{c}_{x}^{\ast}]_{r_{0},1}>[\mathbf{c}_{-x}^{\ast}]_{1,1}$. $\mathbf{D}_{x}$ consists of the intra-subject data points of $\mathbf{x}$ and $\mathbf{D}_{-x}$ consists of the inter-subject data points of $\mathbf{x}$. $[\mathbf{c}_{x}^\ast]_{r_{x},1}$ denotes the $r_{x}$-th largest absolute value of the entries of $\mathbf{c}_{x}^\ast$, and $r_{x}$ is the dimensionality of $\mathcal{S}_{\mathbf{D}}$. Note that, noise and outlier could be regarded as a kind of inter-subject data point of $\mathbf{x}$.
\end{lemma}

\begin{lemma}[\cite{peng2015robust}]
\label{lem2}
Consider a  nonzero data point $\mathbf{x}$ in the intersection between $\mathcal{S}_{\mathbf{D}_{x}}$ and $\mathcal{S}_{\mathbf{D}_{-x}}$, i.e., $\mathbf{x}\in \{\mathcal{S}|\mathcal{S}=\mathcal{S}_{\mathbf{D}_{x}}\cap \mathcal{S}_{\mathbf{D}_{-x}}\}$. Let $\mathbf{c}^{\ast}$, $\mathbf{z}_{0}$, and $\mathbf{z}_{e}$ be the optimal solution of
\begin{equation}
\label{lem2equ1}
\min\hspace{1mm}\|\mathbf{c}\|_{2} \hspace{3mm} \mathrm{s.t.}\hspace{1mm} \mathbf{x}=\mathbf{D}\mathbf{c}
\end{equation}
over $\mathbf{D}$, $\mathbf{D}_{x}$, and $\mathbf{D}_{-x}$, and $\mathbf{c}^{\ast}=\begin{bmatrix}\mathbf{c}_{x}^{\ast}\\\mathbf{c}_{-x}^{\ast} \end{bmatrix}$ is partitioned according to the sets $\mathbf{D}=[\mathbf{D}_{x}\ \mathbf{D}_{-x}]$. If $\|\mathbf{z}_{0}\|_p < \|\mathbf{z}_{e}\|_p$, then $\mathbf{c}_{x}^{\ast}\ne \mathbf{0}$ and $\mathbf{c}_{-x}^{\ast}=\mathbf{0}$.
\end{lemma}

Based on Lemmas~\ref{lem1} and \ref{lem2}, we can obtain a robust representation by truncating the coefficients over errors. Mathematically, 

\begin{equation}
	\label{eq3.4}
	\mathbf{Z}=\mathcal{H}_{\epsilon}(\mathbf{C}^{\ast}), 
\end{equation}
where the hard thresholding operator $\mathcal{H}_{\epsilon}(\mathbf{C})$ keeps large entries and eliminates trivial ones for each column of $\mathbf{C}^{\ast}$. $\mathbf{C}^{\ast}$ is the optimal solution of eq.\ref{eq3.3} which is also the minimizer of eq.\ref{lem1eq1}. 

\begin{figure*}[!t]
\centering{
\subfigure[]{\label{fig3.a}\includegraphics[width=0.48\textwidth]{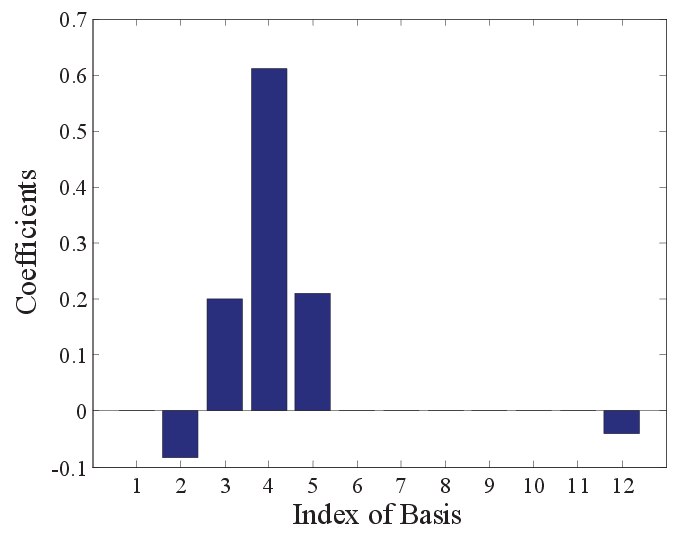}}\hspace{1mm}
\subfigure[]{\label{fig3.b}\includegraphics[width=0.48\textwidth]{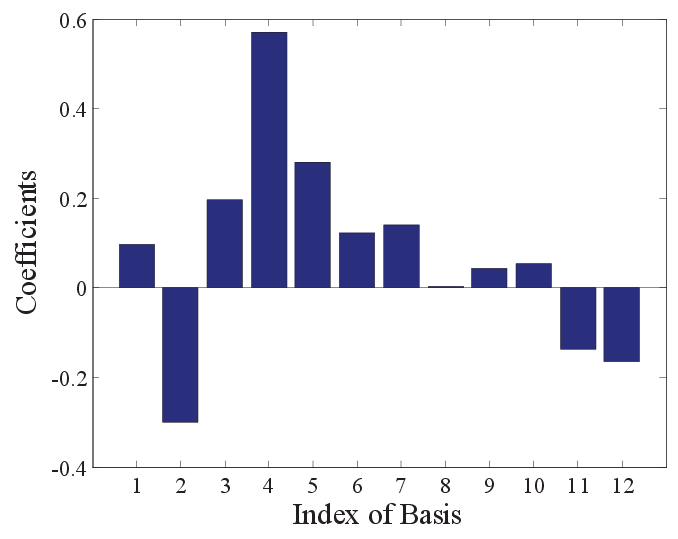}}\\
\subfigure[]{\label{fig3.c}\includegraphics[width=0.48\textwidth]{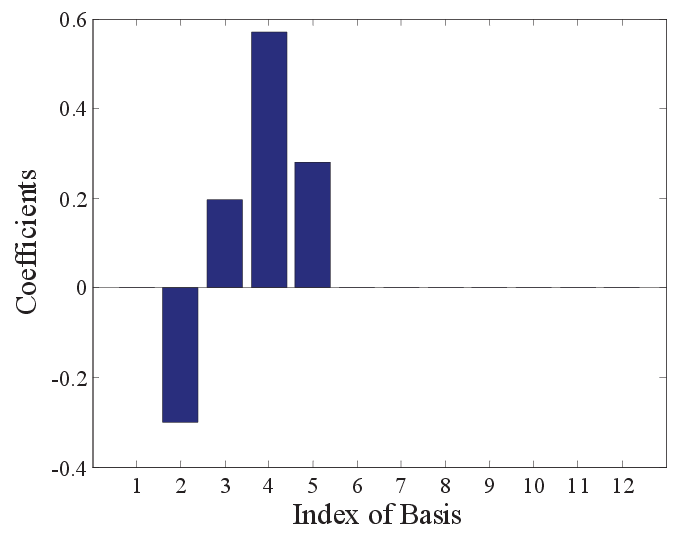}}\hspace{1mm}
\subfigure[]{\label{fig3.d}\includegraphics[width=0.49\textwidth]{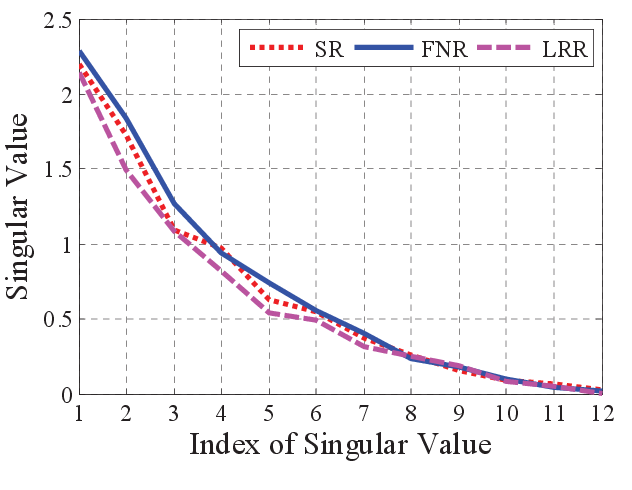}}
}
\caption{An example using two subsets of Extended Yale database B, where the testing subset and the dictionary subset consist of 12 samples, respectively. The first and the last six atoms of the dictionary belong to two different subjects. (a)--(c) sparse representation (eq.\ref{eq2.3}), $\ell_2$-regularized representation (eq.\ref{eq3.3}), and the proposed LrrSPM (eq.\ref{eq3.4}) of a given querying sample that belongs to the first subject. (d) The singular value of the coefficient matrices.~\figurename~\ref{fig3.a}--\ref{fig3.c} show that only the coefficients of LRR over the second subject are zeroes. This makes our model more discriminative. Moreover,~\figurename~\ref{fig3.d} shows that the energy of our method is more concentrated, i.e., our model is more competitive in terms of the principle of minimum description length.}
\label{fig3}
\end{figure*}

\figurename~\ref{fig3} shows a comparison among sparse code (eq.\ref{eq2.3}), $\ell_2$-norm regularized representation (eq.\ref{eq3.3}), and LrrSPM (eq.\ref{eq3.4}). In this example, we carry out  experiments using two subsets of Extended Yale database B~\cite{Georghiades2001}, where the dictionary subset and the testing subset consists of 12 sample, respectively. We randomly select one sample from the first subject as testing sample and calculated its representation.~\figurename~\ref{fig3.a}--\ref{fig3.c} illustrates the obtained representations and \figurename~\ref{fig3.d} shows the singular values of the representation matrix for all testing data. From the results, the proposed method has the following advantages: 1) LrrSPM is more discriminative since its coefficients over the second subjects are zeroes. 2) it provides a compact representation with better  representative capacity.

\begin{algorithm}[t]
\begin{small}
    \caption{Fast LRR for Spatial Pyramid Matching (LrrSPM).}
    \label{alg1}
    \begin{algorithmic}[1]
    \REQUIRE
    The codebook $\mathbf{D}\in \mathds{R}^{m\times k}$, the input image $\mathbf{y}$, and the regularization parameter $\lambda$.
        \STATE Calculate $\mathbf{P}=\left(\mathbf{D}^{T}\mathbf{D} + \lambda \mathbf{I}\right)^{-1}\mathbf{D}^{T}$ and store it.
   \STATE For each image $\mathbf{y}$, detect and extract the SIFT descriptors $\mathbf{X}$ from $\mathbf{y}$.       
    \STATE Calculate the representation of $\mathbf{y}$ via $\mathbf{C} = \mathbf{P}\mathbf{X}$ and normalize each column of $\mathbf{C}$ to have a unit $\ell_2$-norm.
    \STATE If the dictionary contains errors, obtain the LRR of $\mathbf{y}$ by thresholding the trivial entires of $\mathbf{c}_{ji}=[c_{1i}, c_{2i}, \cdots, c_{ki}]^{T}$ at $\epsilon$ (generally, $\epsilon=98\%$) via
   \begin{equation}
    c_{ji}=\left\{
    \begin{aligned} 
    c_{ji} &\hspace{1cm}  k\frac{c_{ji}}{\sum_{j}c_{ji}}<\epsilon\\
    0      &\hspace{1cm} otherwise
    \end{aligned}
    \right.
    \end{equation}   
    \STATE Divide $\mathbf{C}$ into $2^{l}\times 2^{l}$ blocks, where $l$ denotes the scale or the level of the pyramid. For each block at each level, perform max pooling for each block at each level via    
    $\mathbf{z}_{i}=\max\{|\mathbf{c}_{i}^{1}|, |\mathbf{c}_{i}^{2}|,\cdots,|\mathbf{c}_{i}^{b}|\}$, 
    where $\mathbf{c}_{i}^{j}$ denotes the $j$-th LRR vector belonging to the $i$-th block, and $b=2^{l}\times 2^{l}$.
    \ENSURE Form a single representation vector for $\mathbf{y}$ by concatenating the set of $\mathbf{z}_{i}$.
    \end{algorithmic}
    \end{small}
\end{algorithm}

Algorithm~\ref{alg1} summarizes our algorithm. Similar to~\cite{Lazebnik2006,Yang2009}, the codebook $\mathbf{D}$ can be generated by the k-means clustering method or dictionary learning methods such as~\cite{Gao2014TIP}. For training or testing purpose, LrrSPM can get the low rank representation in an online way, which further explores the potential of LRR in online and incremental learning. Moreover, our method is very efficient since its coding process only involves a simple projection operation.

\section{Experiments}
\label{sec4}

\subsection{Baseline Algorithms and Databases}
\label{sec4.1}

We compared our method with four SPM methods using nine image databases. The MATLAB code of LrrSPM can be downloaded from the authors' website \textcolor{blue}{\url{www.machineilab.org/users/pengxi}} and the codes of the baseline methods are publicly accessible. Besides our own experimental results, we also quote some results in the literature.

The baseline methods include BOF~\cite{Fei2005} with linear SVM (LinearBOF) and kernel SVM (KernelBOF), SPM~\cite{Lazebnik2006} with linear SVM (LinearSPM) and kernel SVM (KernelSPM), Sparse Coding based SPM with linear SVM(ScSPM)~\citep{Yang2009}, and Locality-constrained Linear Coding with linear SVM (LLC)~\cite{Wang2010}. 

The used databases include five scene image data sets, three object image data sets (i.e., 17flowers~\cite{Nilsback06}, COIL20~\cite{COIL20} and COIL100~\cite{COIL100}), and one facial image database (i.e., Extended Yale B~\cite{Georghiades2001}). The scene image data sets are from Oliva and Torralba~\cite{Oliva2001}, Fei-Fei and Perona~\cite{Fei2005}, Lazebnik et\ al.~\cite{Lazebnik2006}, Fei-Fei et\ al.~\cite{Fei2006}, and Griffin et\ al.~\cite{Caltech256} which are referred to as OT, FP, LS,  Caltech101, and Caltech256, respectively. \tablename~\ref{tab2} gives a brief review on these data sets.

\begin{table}
\begin{center}
\begin{scriptsize}
   \caption{A summarization of the evaluated databases. $s$ denotes the number of classes and $p$ denotes the number of samples for each subject.}
    \label{tab2}
    \begin{tabular}{lccrrr}
    \toprule
\multicolumn{1}{l}{Databases} & \multicolumn{1}{c}{Type} & \multicolumn{1}{c}{Data Size} & \multicolumn{1}{c}{Image Size} & \multicolumn{1}{c}{$p$} & \multicolumn{1}{c}{$s$} \\
    \midrule
   OT              & scene      &     2688  &  $256\times 256$ & 260--410 & 8\\
   FP                & scene    &     3860  &  $250\times 300$ & 210--410 & 13\\
   LS        & scene           &     4486  &  $250\times 300$ & 210--410 & 15\\
   Caltech101    & scene  &     9144  &   $300\times 200$        & 31--800 & 102\\
   Caltech256   & scene  &     30,607  &   $300\times 200$        & 80--827 & 256\\
   17flowers    & flowers  &     1,360  &   $-$        & 80 & 17\\
   COIL20          & object  &     1440  &  $128\times 128$  & 72 & 20\\
   COIL100       & object   &     7200  &  $128\times128$  & 72 & 100\\
   Extended Yale B  & face &  2414  &  $168\times192$  & 59--64 & 38\\
    \bottomrule
    \end{tabular}
\end{scriptsize}
\end{center}
\end{table}

\subsection{Experimental setup}
\label{sec4.2}

To be consistent with the existing works~\cite{Lazebnik2006,Wang2010,Yang2009}, we use dense sampling technique to divide each image into $2^{l}\times 2^{l}$ blocks (patches) with a step size of 6 pixels, where $l=0,1,2$ denotes the scale. And we extract the SIFT descriptors from each block as features. To obtain the codebook, we use the k-means clustering algorithm to find 256 cluster centers for each data set and use the same codebook for different algorithms. In each test, we split the samples per subject into two parts, one is for training and the other is for testing. Following the common benchmarking procedures, we repeat each experiment five times with different training and testing data partitions and record the average of per-subject recognition rates and the time costs for each test. We report the final results by the mean and standard deviation of the recognition rates and the time costs. For the LrrSPM approach, we fix $\epsilon=0.98$ and assign different $\lambda$ for different databases. For the competing approaches, we directly adopt the parameters configuration in the original works~\cite{Lazebnik2006, Wang2010, Yang2009}. Moreover, we also quote the performance of these methods reported in the original works.

\subsection{Influence of the parameters}
\label{sec4.3}

LrrSPM has two user-specified parameters, the regularization parameter $\lambda$ is used to avoid overfitting and the thresholding parameter $\epsilon$ is used to eliminate the effect of the errors. In this section, we investigate the influence of these two parameters on OT data set. We fix $\epsilon=0.98$ ($\lambda=0.7$) and reported the mean classification accuracy of LrrSPM with the varying $\lambda$ ($\epsilon$). \figurename~\ref{fig3} shows the results, from which one can see that LrrSPM is robust to the choice of the parameters. When $\lambda$ increases from 0.2 to 2.0 with an interval of 0.1, the accuracy ranges from 83.68\% to 85.63\%; When $\epsilon$ increases from 50\% to 100\% with an interval of 2\%, the accuracy ranges from 84.07\% to 86.03\%.

\begin{figure*}[!t]
\centering{
\subfigure[The influence of $\lambda$, where $\epsilon=0.98$.]{\label{fig4.a}\includegraphics[width=0.50\textwidth]{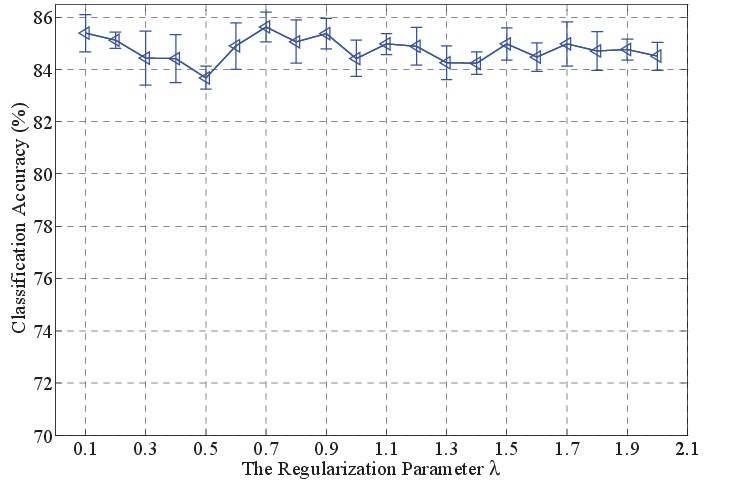}}
\subfigure[The influence of $\epsilon$, where $\lambda=0.70$.]{\label{fig4.b}\includegraphics[width=0.48\textwidth]{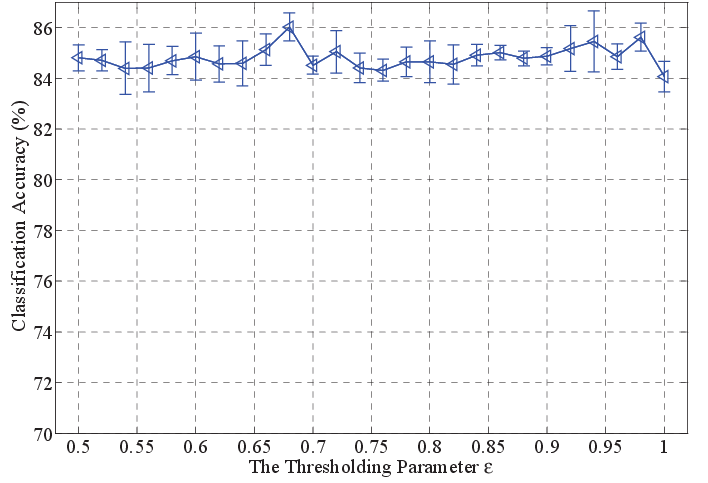}}
}
\caption{The mean and standard deviation of the recognition rates of LrrSPM on the OT database. }
\label{fig4}
\end{figure*}

\subsection{Performance with Different Sized Codebooks}
\label{sec4.4a}

In this Section, we evaluate the performance of LrrSPM when the size of codebook increases from $256$ to $4096$. we carry out experiments on the Caltech101 data set by randomly selecting 30 samples per subject for training and using the rest for testing. The $\lambda$ is set as $0.7$ for LrrSPM. \tablename~\ref{tab3} shows that with increasing $k$, LrrSPM achieves better recognition results but takes more time for coding and classification. Specifically, when $k$ increases from 256 to 4096, the accuracy increases by 4.48\%, but the computing time increases by 1397\%.

\begin{table*}[t]
\caption{The performance of LrrSPM on Caltech101 w.r.t. increasing size of codebook, where the size of codebook $k$ increases from $256$ to $4096$. }
\label{tab3}
\begin{center}
\begin{scriptsize}
\begin{tabular}{l| ccccc}
\toprule
Metric	& $k=256$ &	$k=512$ &  $k=1024$ &   $k=2048$ &  $k=4096$ \\
\midrule
Accuracy	& $65.89\pm1.03$	& $67.75\pm0.92$	& $68.43\pm1.04$	& $69.81\pm1.01$	& $70.37\pm0.98$\\
Time cost & $640.83$ 	& $1005.74$	& $1691.35$	& $3697.03$	& $8952.43$\\
\bottomrule
\end{tabular}
\end{scriptsize}
\end{center}
\end{table*}

\subsection{Scene Classification}
\label{sec4.4}

In this section, the experimental studies consist of two parts. The first part reports the performance of LrrSPM on three scene image databases. The codebook consists of $256$ bases identifying by the k-means method. For each data set, we randomly choose 100 samples from each subject for training and used the rest for testing.

\begin{table*}[t]
\caption{The classification accuracy and the time cost of different methods on the OT, FP, and LS databases.}
\label{tab4}
\begin{center}
\begin{scriptsize}
\begin{tabular}{l| lr| lr| lr}
\toprule
\multicolumn{1}{c|}{\multirow{2}{*}{Algorithms}} & \multicolumn{2}{c|}{\multirow{1}{*}{the OT database}} & \multicolumn{2}{c|}{\multirow{1}{*}{the FP database}} & \multicolumn{2}{c}{the LS database}\\
\cline{2-7}
& \multicolumn{1}{c}{Accuracy (\%)}& \multicolumn{1}{c|}{Time (s)}& \multicolumn{1}{c}{Accuracy (\%)} & \multicolumn{1}{c|}{Time (s)}& \multicolumn{1}{c}{Accuracy (\%)}& \multicolumn{1}{c}{Time (s)}\\
\midrule
LrrSPM (Ours)        & \textbf{85.63}$\pm$\textbf{0.56} & $116.67  $ & \textbf{80.90}$\pm$\textbf{0.75} & $159.84  $ & \textbf{76.34}$\pm$\textbf{0.58} & $189.79 $\\
LinearBOF     & $78.95\pm0.12$ & $102.76  $ & $72.13\pm0.43$ & $195.85  $ & $66.38\pm0.88$ & $169.25 $\\
KernelBOF     & $76.28\pm0.16$ & $103.65  $ & $68.18\pm0.24$ & $203.89  $ & $62.67\pm0.78$ & $178.34 $\\
linearSPM     & $76.09\pm0.55$ & $209.14  $ & $71.63\pm0.51$ & $239.44  $ & $66.91\pm0.78$ & $188.24 $\\
KernelSPM     & $73.52\pm0.64$ & $196.20  $ & $64.97\pm1.50$ & $266.96  $ & $60.83\pm0.39$ & $196.35 $\\
ScSPM & $84.44\pm0.24$ & $5375.41 $ & $79.04\pm0.91$ & $5841.89 $ & $74.40\pm0.45$ & $9539.62$\\ 
LLC           & $85.55\pm0.34$ & $640.27  $ & $80.34\pm0.76$ & $943.81  $ & \textbf{76.99}$\pm$\textbf{1.21} & $1059.96$\\ 
\hline
Rasiwasia's method & \multicolumn{1}{c}{-} & \multicolumn{1}{c|}{-} &\multicolumn{1}{l}{76.20} & \multicolumn{1}{c|}{-} & $72.50\pm 0.30$& \multicolumn{1}{c}{-}\\
\bottomrule
\end{tabular}
\end{scriptsize}
\end{center}
\end{table*}

\tablename~\ref{tab4} shows that LrrSPM is slightly better than the other evaluated algorithms in most tests. Although LrrSPM is not the fastest method, it finds a good balance between the efficiency and the classification rate. On the OT database, the speed of LrrSPM is about 5.49 and 46.07 times faster than ScSPM and LLC, respectively. On the LS database, the speedups are 5.59 and 50.26 times.

The second part of experiment reports the performance of the evaluated methods using Caltech101, Caltech256, and Oxford 17flowers database by randomly selecting 30 samples per subject for training and using the rest for testing. In the tests, the dictionary contains 256 bases identified by the k-means clustering method. We fix $\lambda=0.7$, $\lambda=0.24$, and $\lambda=0.145$ for LrrSPM on these three data sets. 

\tablename~\ref{tab5} reports the results from which we can find that, on the Caltech101 data set, the recognition rates of LrrSPM is 28.78\% higher than that of LinearBOF, 20.73\% higher than that of Kernel BOF, 22.36\% higher than that of LinearSPM, 12.38\% higher than that of KernelSPM, 0.5\% higher than that of ScSPM and $1.97\%$ lower than that of LLC. However, LrrSPM only takes about 3\% (30\%) CPU time of ScSPM (LLC). On the 17flowers database, LrrSPM outperforms the other evaluated methods by a considerable performance margin. Its recognition rate is 4.98\% higher than ScSPM with 21 times speedup.

\begin{table*}[t]
\caption{The classification accuracy and the time cost of different methods on the Caltech101, Caltech256, and  17flowers database.}
\label{tab5}
\begin{center}
\begin{scriptsize}
\begin{tabular}{l| lr| lr| lr}
\toprule
\multicolumn{1}{c|}{\multirow{2}{*}{Algorithms}} & \multicolumn{2}{c|}{\multirow{1}{*}{Caltech101}} & \multicolumn{2}{c|}{\multirow{1}{*}{Caltech256}} & \multicolumn{2}{c}{17flowers}\\
\cline{2-7}
& \multicolumn{1}{c}{Accuracy (\%)}& \multicolumn{1}{c|}{Time (s)}& \multicolumn{1}{c}{Accuracy (\%)} & \multicolumn{1}{c|}{Time (s)}& \multicolumn{1}{c}{Accuracy (\%)}& \multicolumn{1}{c}{Time (s)}\\
\midrule
LrrSPM (Ours) & 65.89$\pm$ 1.03 & 640.83  & 27.43$\pm$ 0.98 & 2518.13 & 61.42$\pm$ 1.46 & 121.00\\
LinearBoF & 37.11$\pm$ 1.11 & 307.50  & 15.09$\pm$ 1.45 & 2474.96 & 40.19$\pm$ 1.90 & 45.79\\
KernelBoF & 45.16$\pm$ 0.81 & 315.62  & 11.25$\pm$ 0.46 & 3035.73 & 34.70$\pm$ 2.79 & 47.81\\
linearSPM & 43.53$\pm$ 1.17 & 410.79  & 23.12$\pm$ 0.27 & 2511.19 & 44.12$\pm$ 2.45 & 86.66\\
KernelSPM & 53.51$\pm$ 1.11 & 467.68  & 12.03$\pm$ 0.48 & 5819.58 & 36.35$\pm$ 2.20 & 94.79\\
ScSPM & 65.39$\pm$ 1.21 & 18964.84  & 28.60$\pm$  0.15 & 58313.03 & 56.44$\pm$  0.54 & 2614.39\\
LLC & 67.86$\pm$  1.17 & 2203.58  & 29.35$\pm$  0.42 & 6893.68 & 59.89$\pm$  1.55 & 747.49\\
\bottomrule
\end{tabular}
\end{scriptsize}
\end{center}
\end{table*}

\begin{table*}[t]
\caption{The  classification accuracy on Caltech101 and Caltech256 reported by some recent literatures. DBN, CNN, and OtC are the abbreviations of deep belief network, convolutional neural network, and object to class method, respectively.}
\label{tab6}
\begin{center}
\begin{scriptsize}
\begin{tabular}{l lll rrr}
\toprule
KernelSPM	&ScSPM &	LLC	& GMatching & DBN	& CNN & OtC\\
\midrule
$64.60\pm0.80$	&$73.20\pm0.54$	&$73.44$	&$80.30\pm1.2$	&$65.40$&	$66.30$	&$64.26$\\
$-$	&$34.02\pm0.35$	&$41.19$	&$38.1$	&$-$&	$-$	&$-$\\
\bottomrule
\end{tabular}
\end{scriptsize}
\end{center}
\end{table*}

Besides our experimental implementations, \tablename~\ref{tab6} summarizes some state-of-the-art results reported by~\cite{Duchenne2011,Lazebnik2006,Yang2009,Lee2009,Kavukcuoglu2010,Zhang2014Learning} on Caltech101. One can find that we do not reproduce the results reported in the literature for some evaluated methods. This could be attributed to the subtle engineering details. Specifically, Lazebnik et\ al.~~\cite{Lazebnik2006} only used a subset of Caltech101 (50 images for each subject) rather than all samples.~\cite{Wang2010,Yang2009} used a larger codebook ($k=2048$) and the codebook could probably be different even though the size of codebook is fixed due to the randomness. Duchenne et\ al.~\cite{Duchenne2011} reported the state-of-the-art accuracy of 80.30\% and 38.1\% on Caltech101 and Caltech256 by using graph matching based method to improve the performance of classifier. With multiple kernel learning, Yang et\ al.~\cite{Yang2009Group}  proposed a model which achieves 84.3\% of  accuracy on Caltech101. This significant improvement may attribute to the nonlinearity of kernel functions. Moreover, Todorovic and Ahuja~\cite{Todorovic2008CVPR} showed that the performance of model can be further improved by employing ensemble learning method over multiple descriptors. Their method achieves 49.5\% of  accuracy on Caltech256 by fusing six different descriptors.

\subsection{Object and Face Recognition}
\label{4.6}

This section investigates the performance of LrrSPM on two object image data sets (i.e., COIL20 and COIL100) and one facial image database (i.e., Extended Yale Database B). To analyze the time costs of the examined methods, we also report the time costs of the methods for encoding and classifying.  

\begin{table*}[t]
\caption{Object image classification results of different methods on the COIL20 database with different training samples.}
\label{tab7}
\begin{center}
\begin{scriptsize}
\begin{tabular}{l| rrrrr}
\toprule
\multicolumn{1}{c|}{\multirow{2}{*}{Algorithms}} & \multicolumn{5}{c}{\multirow{1}{*}{Training Images for Each Subject}}\\
\cline{2-6}
& \multicolumn{1}{c}{10} &  \multicolumn{1}{c}{20} & \multicolumn{1}{c}{30} & \multicolumn{1}{c}{40} & \multicolumn{1}{c}{50} \\
\midrule
LrrSPM (Ours)        & \textbf{97.90}$\pm$\textbf{0.42} & \textbf{99.52}$\pm$\textbf{0.87} & \textbf{100.00}$\pm$\textbf{0.00} & \textbf{100.00}$\pm$\textbf{0.00} & \textbf{100.00}$\pm$\textbf{0.00} \\      
LinearBOF    & $87.55\pm0.17$ & $94.08\pm0.94$ & $96.65\pm0.42$  & $97.38\pm0.26$  & $98.46\pm0.84$  \\    
KernelBOF    & $86.47\pm0.27$ & $95.60\pm1.62$ & $97.43\pm1.10$  & $98.41\pm1.24$  & $98.46\pm1.09$  \\    
linearSPM    & $85.00\pm1.00$ & $92.23\pm1.85$ & $94.17\pm1.39$  & $97.09\pm1.17$  & $97.09\pm0.57$  \\    
KernelSPM    & $86.61\pm0.76$ & $93.14\pm2.30$ & $95.41\pm0.63$  & $98.28\pm1.56$  & $98.64\pm1.41$  \\    
ScSPM         & $97.09\pm1.13$ & $98.85\pm0.98$ & $99.65\pm0.76$  & \textbf{100.00}$\pm$\textbf{0.00} & \textbf{100.00}$\pm$\textbf{0.00} \\
LLC          & $97.17\pm1.14$ & $99.32\pm1.00$ & $99.64\pm0.89$  & $99.84\pm0.89$  & \textbf{100.00}$\pm$\textbf{0.00} \\              
\bottomrule
\end{tabular}
\end{scriptsize}
\end{center}
\end{table*}

\begin{table*}[t]
\caption{Object image classification results of different methods on the COIL100 database with different training samples.}
\label{tab8}
\begin{center}
\begin{scriptsize}
\begin{tabular}{l| rrrrr}
\toprule
\multicolumn{1}{c|}{\multirow{2}{*}{Algorithms}} & \multicolumn{5}{c}{\multirow{1}{*}{Training Images for Each Subject}}\\
\cline{2-6}
& \multicolumn{1}{c}{10} &  \multicolumn{1}{c}{20} & \multicolumn{1}{c}{30} & \multicolumn{1}{c}{40} & \multicolumn{1}{c}{50} \\
\midrule
LrrSPM (Ours)      & \textbf{91.19}$\pm$\textbf{0.65} & \textbf{97.39}$\pm$\textbf{0.78}& \textbf{99.29}$\pm$\textbf{0.21} & \textbf{99.87}$\pm$\textbf{0.36} & \textbf{99.85}$\pm$\textbf{0.07}\\
LinearBOF  & $84.32\pm1.15$ & $91.65\pm0.32$ & $94.76\pm0.35$ & $95.99\pm0.42$ & $96.81\pm0.36 $\\
KernelBOF  & $82.32\pm1.12$ & $92.77\pm0.53$ & $94.01\pm0.75$ & $96.36\pm0.66$ & $97.20\pm0.48 $\\
linearSPM  & $84.84\pm0.64$ & $92.17\pm0.63$ & $95.30\pm0.53$ & $96.46\pm0.28$ & $97.64\pm0.33 $\\
KernelSPM  & $86.01\pm0.12$ & $92.62\pm0.61$ & $96.49\pm0.98$ & $97.56\pm0.88$ & $98.29\pm0.32 $\\
ScSPM      & $90.56\pm0.34$ & $94.73\pm0.57$ & $97.62\pm0.15$ & $98.44\pm0.10$ & $99.81\pm0.07 $\\
LLC        & \textbf{91.26}$\pm$\textbf{0.42} & $96.35\pm0.65$ & $97.97\pm0.34$ & $98.49\pm0.24$ & $99.81\pm0.19 $\\
\bottomrule
\end{tabular}
\end{scriptsize}
\end{center}
\end{table*}

\begin{table*}[t]
\caption{Face image classification results of different methods on the Extended YaleB Database B with different training samples.}
\label{tab9}
\begin{center}
\begin{scriptsize}
\begin{tabular}{l| rrrrr}
\toprule
\multicolumn{1}{c|}{\multirow{2}{*}{Algorithms}} & \multicolumn{5}{c}{\multirow{1}{*}{Training Images for Each Subject}}\\
\cline{2-6}
& \multicolumn{1}{c}{10} &  \multicolumn{1}{c}{20} & \multicolumn{1}{c}{30} & \multicolumn{1}{c}{40} & \multicolumn{1}{c}{50} \\
\midrule
LrrSPM (Ours)     & \textbf{87.08}$\pm$\textbf{0.41} & \textbf{96.03}$\pm$\textbf{0.89} & \textbf{98.28}$\pm$\textbf{0.55} & \textbf{99.23}$\pm$\textbf{0.81} & \textbf{99.81}$\pm$\textbf{0.83}\\
LinearBOF  & $50.26\pm1.25$ & $64.13\pm0.73$ & $70.66\pm1.28$ & $73.78\pm0.60$ & $77.21\pm2.14$\\
KernelBOF  & $59.59\pm2.33$ & $63.51\pm1.27$ & $71.35\pm1.40$ & $76.25\pm1.67$ & $83.56\pm1.98$\\
linearSPM  & $50.21\pm3.60$ & $70.05\pm1.20$ & $80.82\pm1.95$ & $84.39\pm0.60$ & $88.68\pm1.73$\\
KernelSPM  & $54.49\pm2.52$ & $71.80\pm1.32$ & $85.54\pm3.53$ & $84.84\pm2.10$ & $88.90\pm3.10$\\
ScSPM      & $86.79\pm0.20$ & $94.22\pm0.45$ & $98.05\pm0.44$ & $99.00\pm0.87$ & $99.57\pm1.16$\\
LLC        & $84.79\pm0.59$ & $95.45\pm0.64$ & $98.05\pm0.41$ & $98.98\pm0.25$ & $99.23\pm0.93$\\   
\bottomrule
\end{tabular}
\end{scriptsize}
\end{center}
\end{table*}

\begin{table*}[t]
\caption{The time costs (seconds) for encoding and classification (including training and testing) of different methods on three image databases. The speed of LrrSPM is 25.67--42.58 times faster than ScSPM and 9.58--16.68 times faster than LLC.}
\label{tab10}
\begin{center}
\begin{scriptsize}
\begin{tabular}{l| rr| rr| rr}
\toprule
\multicolumn{1}{c|}{\multirow{2}{*}{Algorithms}} & \multicolumn{2}{c|}{\multirow{1}{*}{COIL20}} & \multicolumn{2}{c|}{\multirow{1}{*}{COIL100}}& \multicolumn{2}{c}{\multirow{1}{*}{Extended Yale B}}\\
\cline{2-7}
& \multicolumn{1}{c}{Coding} &  \multicolumn{1}{c|}{Classification} & \multicolumn{1}{c}{Coding} &  \multicolumn{1}{c|}{Classification}& \multicolumn{1}{c}{Coding} &  \multicolumn{1}{c}{Classification} \\
\midrule
LrrSPM     & $16.54$  & $1.4$   & $43.91$   & $4.49$ & $49.67$   & $2.69$\\
LinearBOF  & $11.54$  & $0.12$  & $11.53$   & $0.11$  & $59.92$   & $0.26$\\
KernelBOF  & $11.54$  & $11.54$ & $11.53$   & $0.78$  & $59.92$   & $2.05$\\
linearSPM  & $12.15$  & $0.17$  & $78.92$   & $2.28$  & $93.38$   & $0.52$\\
KernelSPM  & $12.15$  & $1.5$   & $78.92$   & $36.79$ & $93.38$   & $8.25$\\
ScSPM      & $424.48$ & $0.79$  & $1837.03$ & $4.07$  & $2114.88$ & $3.08$\\
LLC        & $275.94$ & $3.86$  & $432.2$   & $6.34$  & $475.94$  & $3.86$\\
\bottomrule
\end{tabular}
\end{scriptsize}
\end{center}
\end{table*}

\tablename s~\ref{tab7}--~\ref{tab9} report the recognition rate of the tested approaches on COIL20, COIL100, and Extended Yale B, respectively. In most cases, our method achieves the best results and is followed by ScSPM and LLC. When 50 samples per subject of  COIL20 and COIL100 are used for training the classifier, LrrSPM groups all the testing images into the correct categories.  On the Extended Yale B, LrrSPM also classifies almost all the samples into the correct categories (the recognition rate is about 99.81\%). 

\tablename~\ref{tab10} shows the efficiency of the evaluated methods. One can find that LrrSPM, BOF, and SPM are more efficient than ScSPM and LLC both in the process of encoding and classification. Specifically, the CPU time of LrrSPM is only about 2.35\%--3.90\% of that of ScSPM and about 5.99\%--10.44\% of that of LLC.

\section{Conclusion}
\label{sec5}

In this paper, we proposed a spatial pyramid matching method which is based on the lowest rank representation (LRR) of the SIFT descriptors. The proposed method, named as LrrSPM, formulates the quantization of the SIFT descriptors as a rank minimization problem and utilizes the multiple-scale representation to characterize the statistical information of the image. LrrSPM is very efficient in computation while still maintaining a competitive accuracy on a range of data sets. In general, LrrSPM is 25--50 times faster than ScSPM and 5--16 times faster than LLC. Experimental results based on several well-known data sets show the good performance of LrrSPM.

Each approach has its own advantages and disadvantages. LrrSPM is based on the low rank assumption of data space. If this assumption is unsatisfied, the performance of our method may be degraded. Moreover, although LrrSPM performs comparable to ScSPM and LLC with significant speedup, its performance can be further improved by referring to the recently-proposed methods. By referring to~\cite{Gao2013TIP}, one can develop the nonlinear LrrSPM by incorporating kernel function into our objective function. By referring to~\cite{Zhou2013}, one can utilize the label information to design supervised LrrSPM. Moreover, the performance of LrrSPM can also be improved by fusing multiple descriptors as~\cite{Todorovic2008CVPR} does.

\section*{Acknowledgement}
The authors would like to thank the anonymous editors ad reviewers for their valuable comments and suggestions to improve the quality of this paper. This work was supported by National Nature Science Foundation of China under grant No.61432012.

\section*{Reference}
\bibliographystyle{elsarticle-num}
\begin{small}
\bibliography{LrrSPM}
\end{small}

\end{document}